\documentclass[conference]{IEEEtran}
\pdfoutput=1
\IEEEoverridecommandlockouts
\usepackage{cite}
\usepackage{amsmath,amssymb,amsfonts}
\usepackage{algorithmic}
\usepackage{graphicx}
\usepackage{textcomp}
\usepackage{xcolor}
\usepackage{multirow}
\usepackage{amsthm}
\usepackage{subfigure}

\def\BibTeX{{\rm B\kern-.05em{\sc i\kern-.025em b}\kern-.08em
    T\kern-.1667em\lower.7ex\hbox{E}\kern-.125emX}}
\begin{document}

\title{Spatio-Temporal Hypergraph Neural ODE Network for Traffic Forecasting}

\author{\IEEEauthorblockN{Chengzhi Yao, Zhi Li, Junbo Wang\textsuperscript{\dag}\thanks{\dag \ is the corresponding author.}}
\IEEEauthorblockA{\textit{School of Intelligent Systems Engineering, Sun Yat-Sen University, China}\\
yaochzh3@mail2.sysu.edu.cn, lizh337@mail2.sysu.edu.cn, wangjb33@mail.sysu.edu.cn\\
}}

\maketitle

\begin{abstract}
Traffic forecasting, which benefits from mobile Internet development and position technologies, plays a critical role in Intelligent Transportation Systems. It helps to implement rich and varied transportation applications and bring convenient transportation services to people based on collected traffic data. Most existing methods usually leverage graph-based deep learning networks to model the complex road network for traffic forecasting shallowly. Despite their effectiveness, these methods are generally limited in fully capturing high-order spatial dependencies caused by road network topology and high-order temporal dependencies caused by traffic dynamics. To tackle the above issues, we focus on the essence of traffic system and propose STHODE: Spatio-Temporal Hypergraph Neural Ordinary Differential Equation Network, which combines road network topology and traffic dynamics to capture high-order spatio-temporal dependencies in traffic data. Technically, STHODE consists of a spatial module and a temporal module. On the one hand, we construct a spatial hypergraph and leverage an adaptive MixHop hypergraph ODE network to capture high-order spatial dependencies. On the other hand, we utilize a temporal hypergraph and employ a hyperedge evolving ODE network to capture high-order temporal dependencies. Finally, we aggregate the outputs of stacked STHODE layers to mutually enhance the prediction performance. Extensive experiments conducted on four real-world traffic datasets demonstrate the superior performance of our proposed model compared to various baselines.
\end{abstract}


\section{Introduction}
Traffic forecasting has raised intensive attention with the increasing spatio-temporal data collected by entities like governments and transportation companies, which contributes to convenient transportation services, including order dispatching, route planning, and ride sharing. 

Numerous efforts have been made to achieve encouraging accuracy in traffic forecasting by addressing spatio-temporal dependencies within data originating from road network topology and traffic dynamics. Traditionally, early works viewed traffic forecasting as a time series problem and addressed it via statistical and machine learning methods\cite{ARIMA, jeong2013supervised}. However, these methods overlooked spatial dependencies, leading to less-than-ideal performance. Recently, graph neural networks(GNNs) and their variants\cite{dowe,DCRNN,D2STGNN,GMAN,STSGCN,STGODE,graphwavenet} have dominated this field, arising from their remarkable ability to capture correlations among nodes. Prominent approaches, exemplified by STGCN-based methods\cite{DCRNN, D2STGNN, GMAN,STSGCN, STGODE,graphwavenet} characterize the road network topology by representing the pair-wise relationships among nodes using simple graphs and model the traffic dynamics as diffusion progress\cite{DCRNN}.

Although the aforementioned approaches have shown encouraging performance, we argue that the simple graph squeezes the complex spatio-temporal dependencies into pair-wise ones, which leads to incomplete modeling of the road network topology and traffic dynamics. Technically, most GNNs-based works typically adopt graph convolution networks over a simple geographic graph constructed with spatial correlations and model the traffic dynamics as a discrete diffusion process. They face limitations in two critical aspects: i) The use of simple pair-wise graphs does not adequately model the complex road network topology. ii) Discrete GCNs are inadequate to model traffic dynamics for effectively capturing the evolution of the traffic system.

To address these issues, we propose Spatio-Temporal Hypergraph Ordinary Differential Equation Network(STHODE) for traffic forecasting. The key idea of STHODE is to leverage hypergraph structure to represent complex spatial correlations and ordinary differential equations (ODEs) to model the evolution of dynamical systems. To achieve this goal effectively, we introduce two modules, i.e. spatial module and temporal module. In the spatial module, we construct a spatial hypergraph and employ an adaptive MixHop hypergraph ODE layer to capture high-order spatial dependencies caused by the road network topology. In the temporal module, we construct a temporal hypergraph and leverage a hyperedge evolving ODE layer to capture high-order temporal dependencies caused by the traffic dynamics. Furthermore, we aggregate the outputs of the stacked STHODE layers, leveraging their mutual interactions to improve prediction performance within a supervised learning framework. We validate the effectiveness of STHODE on four real-world datasets and the extensive experimental results demonstrate that our STHODE model outperforms various baseline models.
In summary, the main contributions of this paper are as follows:
\begin{itemize}
     \item We propose a spatial module and a temporal module to model the road network topology and traffic dynamics respectively. We construct two types of hypergraphs to enhance the capture of spatio-temporal dependencies.
     \item We present Spatio-Temporal Hypergraph Ordinary Differential Equation Network(STHODE) for traffic forecasting, which addresses the limitations of GNNs-based approaches in modeling road network topology and traffic dynamics. Our proposed method provides improved interpretability compared to existing approaches for traffic forecasting.
    \item We evaluate STHODE on four real-world datasets through extensive experiments, demonstrating its superiority in traffic forecasting compared to various baselines.
\end{itemize}

\section{RELATED WORKS}
In this section, we briefly review the related works in three aspects: Traffic Forecasting, Hypergraph Learning, and Neural Ordinary Differential Equations.

\textbf{Traffic Forecasting.} Spatio-Temporal Graph Neural Networks(STGNNs)\cite{DCRNN, D2STGNN, GMAN, STSGCN, STGODE,graphwavenet} are the most representative approaches to capture spatio-temporal dependencies in traffic forecasting. Most of these methods only consider the pair-wise relationship between traffic sensors and model the traffic dynamics as diffusion progress. Only D2STGNN\cite{D2STGNN} introduces a framework that separates diffusion and the inherent traffic signal, enabling the modeling of traffic dynamics beyond the diffusion process. However, all of them fail to fully model road network topology due to the limitation of simple graphs.

\textbf{Hypergraph learning.} In many real-world problems, relationships among objects are more complex than pair-wise. Hypergraph learning has been employed in various domains to model high-order correlations among data. \cite{NIPS2006_dff8e9c2} first introduced hypergraph learning which conducts transductive learning as a propagation process on the hypergraph in the classification task. With the development of deep learning, HGNN\cite{hypergraphNN} introduced the hypergraph deep learning neural network for data representation learning. Recently, Hypergraph learning has attracted more attention in spatio-temporal prediction. ST-HSL\cite{li2022spatial} unifies hypergraph dependency modeling with self-supervision learning for spatio-temporal crime representations. All these works highlight the remarkable capability of hypergraph learning in capturing high-order correlations among data.

\textbf{Neural Ordinary Differential Equations.} Neural ODE\cite{neuralODE} introduces a novel paradigm for extending discrete deep neural networks to continuous scenarios. CGNN\cite{continuousGNN} extends ODE to graph-structured data. Due to the superior performance and flexible capability, graph ODEs have gained widespread adoption in various research fields, such as traffic forecasting\cite{STGODE}, recommendation, and dynamic interacting systems. STGODE\cite{STGODE} utilizes graph ODE to address the over-smoothing problem and effectively model long-range spatio-temporal dependencies in traffic forecasting. However, the limitations of STGODE lie in pair-wise modeling, so we propose a novel approach that leverages hypergraph ODE for a more comprehensive representation.

\section{Preliminaries}
\subsection{Hypergraph Learning} 
\textbf{Notation 1:(Hypergraph)} Let $G = (V, \xi, H, W, E)$ denotes a hypergraph, with the node set $V = \{v_1,\dots,v_N\}$ and hyperedge set $\xi = \{ e_{1},\dots,e_M\}$. The incidence matrix $H \in \mathbb{R}^{N \times M}$ depicts the connections between nodes and hyperedges, with entries defined as:
\begin{equation}
H_{ij} = 
\begin{cases}
1, & \text{if } v_i \in e_j \\
0, & \text{otherwise}
\end{cases}.     
 \end{equation}

Each hyperedge is assigned with a positive weight $W(e)$ and a positive embedding $E(e)$, with all the weights and embeddings stored in $W = \text{diag}(w_1,\dots,w_M) $ and $E \in \mathbb{R}^{N \times M}$ respectively.

\textbf{Notation 2:(Hypergraph Convolution)} Convolution operator on the hypergraph $G$ is defined based on two assumptions\cite{hypergraph}: 1) More propagation should occur between nodes connected by a hyperedge. 2) Hyperedges with larger weights should have a higher impact on propagation. The hypergraph convolution layer is defined as:
\begin{equation}
X_i^{l+1} = \sigma(\sum_{j=1}^{N} \sum_{m=1}^{M} H_{im} H_{jm} W_{mm} X_j^{l} \mathbf{P}),
\end{equation}
where $X_{i}^{l}$ denotes the embedding of node $v_i$ in the $l$-th layer.$\sigma$ denotes an element-wise activation function. $\mathbf{P} \in \mathbb{R}^{F^{(l)} \times F^{(l+1)}}$ denotes the transform matrix between $l$-th layer and $(l+1)$-th layer. 

\subsection{Neural Ordinary Differential Equations}
We can model the evolution of states in a dynamical system using a first-order ODE, i.e. $\dot{z(t)}:=\frac{dz(t)}{dt}=f(z(t),t)$, where the ODE function $f$ can be parameterized by a neural network. Given the ODE function $f$, the whole trajectory of the object is determined by the initial state $z_0$ as follows:
\begin{equation}
\mathbf{z}(t) = \mathbf{z}_0 + \int_{0}^{T}\frac{dz}{ds}ds = \mathbf{z}_0 + \int_{0}^{T}f(\mathbf{z}(s),s)ds.
\end{equation}
And we can rely on various numerical methods to solve the integral problem, such as Euler and Runge-Kutta.

\subsection{Problem Definition}
\textbf{Notation 3:(Traffic Sensor)} A traffic sensor is a sensor deployed in the road network, which samples traffic signals $\mathbf{X}$ such as flow and vehicle speed.

\textbf{Notation 4:(Road Network)} A road network is represented as a hypergraph $G=(V,\xi,H)$, consisting of different road segments that vary in structure and functionality. The node set $V$ corresponds to traffic sensors, and the hyperedge set $\xi$ corresponds to road segments. The incidence matrix $H$ stores the connection information between traffic sensors and road segments.

\textbf{Problem Statement.} Given the historical traffic signals $\mathbf{X}^{t-T+1:t}=[X_{t-T+1},\dots, X_{t}] \in \mathbb{R}^{T \times N \times F}$ for a sequence of $T$ time steps, we aim to learn a mapping function $f$ that predicts the future traffic signals $\mathbf{X}^{t+1:t+S}=[X_{t+1}, \dots, X_{t+S}] \in \mathbb{R}^{S \times N \times F}$ for the next sequence of $S$ time steps.
\begin{figure*}
  \centering
  \includegraphics[width=0.9\textwidth]{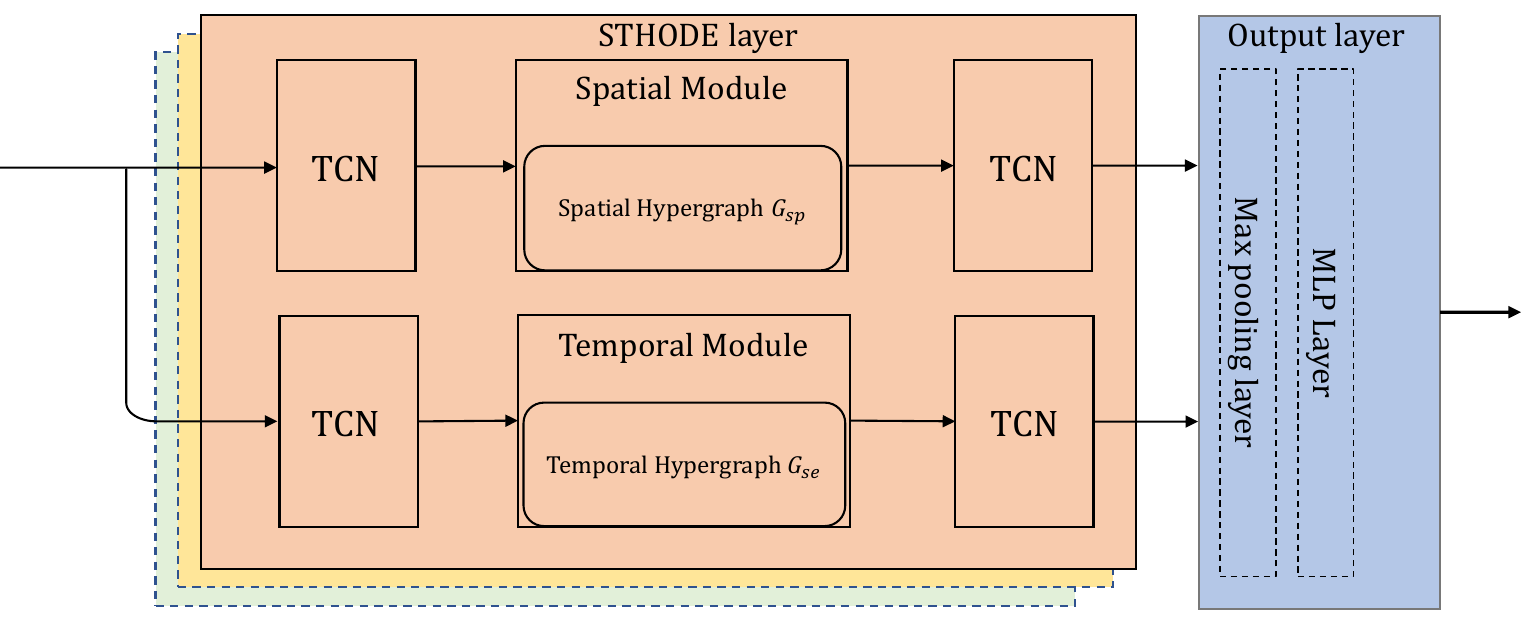}
  \caption{The Framework of STHODE.}
  \label{fig:model}
\end{figure*}

\section{Methodology}
The overall framework of STHODE is shown in Figure \ref{fig:model}. In the following subsections, we introduce how we capture high-order spatial dependencies with the spatial module and high-order temporal dependencies with the temporal module. Furthermore, we aggregate the outputs of the stacked STHODE layers, leveraging their mutual interactions to improve prediction performance within a supervised learning framework.

\subsection{The Spatial Module}\label{sec:spa_mo}
\textbf{1) Construction of Spatial Hypergraph}: We define $G_{sp}=(V, \xi, H, W, E)$ as a spatial hypergraph, with $V$ representing traffic sensors and $\xi$ denoting hyperedges set. The binary incidence matrix $H$ encodes node-hyperedge relationships. To model topological influences, we use diagonal matrix $W$ for road segment types and combine it with hyperedge embedding matrix $E$ to represent road segment impacts on traffic sensors.

In practice, we often lack complete prior knowledge such as road segment information. Following STGCN\cite{STGCN}, we construct a graph adjacency matrix $A^{ng}$ based on node connectivity and distance. Hyperedges are formed using a centroid-based approach from $A^{ng}$, including the centroid and its first and second-order neighbors within radius $R$. The number of nodes in each hyperedge determines its weight $W(e)$. Further, we propose an adaptive hypergraph incidence matrix $\Tilde{H}$ learned end-to-end, enabling adaptively modeling road network topology.
 
Given nodes embedding vector and hyperedges embedding vector with learnable parameters $E_n \in \mathbb{R}^N$ and $E_{m} \in \mathbb{R}^M$, the adaptive hypergraph incidence matrix $\Tilde{H}$ is defined as:
\begin{equation}
    \Tilde{H}=H\odot\text{Softmax}(E_n \otimes E_m^T).
    \label{h}
\end{equation} 
where $\odot$ denotes the Hadamard product and $\otimes$ denotes the Kronecker product. The construction of spatial hypergraph $G_{sp}$ addresses the limitations of incomplete prior knowledge and enables adaptive modeling of road network topology.

\textbf{2) Adaptive MixHop Hypergraph ODE}:
Technically, given the spatial hypergraph $G_{sp}$, we have the normalized adaptive hypergraph matrix $\Tilde{A}$ as follows:
\begin{equation}
    \Tilde{A} = \Tilde{D}^{-\frac{1}{2}}\Tilde{H}\Tilde{B}^{-1}W\Tilde{H} \Tilde{D}^{-\frac{1}{2}},
\end{equation}
where $\Tilde{D} \in \mathrm{R}^{N \times N}$ and $\Tilde{B} \in \mathrm{R}^{M \times M}$ are both diagonal matrices. The diagonal entry $d(v)=\sum_{e \in \xi} W(e)H(v,e)$ denotes the degree of node $v$ and $b(e)=\sum_{v \in V}H(v,e)$ denote the degree of the hyperedge $e$. 
 
Inspired by MixHop GCNs\cite{pmlr-v97-abu-el-haija19a}, we propose the adaptive MixHop hypergraph convolution layers to capture higher-order information locally and globally following the scheme:
\begin{equation}
\label{eq:di}
\textbf{X}_{n+1} = \sum_{k=1}^{K}\Tilde{A}_k \times_1 \mathbf{X}_{n} \times_2 U_k \times_3 Q_k + \textbf{X}_0.
\end{equation}

In the given equation, $\mathbf{X}_n \in \mathrm{N \times T \times F}$ represents hidden representation of nodes in the $n$-th layer. The hyperparameter $K$ is the depth of propagation and $\Tilde{A}_k$ denotes the matrix $\Tilde{A}$ multiplied by itself $k$ times. $U_k \in \mathrm{R}^{T \times T}$ and $Q_k \in \mathrm{R}^{F \times F}$ are both trainable transform matrices. The restart distribution\cite{continuousGNN} $\textbf{X}_0$ denotes the initial input of the propagation layer, which mitigates the issue of information loss and over-smoothing problem. 

However, in the traffic dynamic system, traffic signals and road network status evolve with continuous-time flow. Extending the discrete propagation scheme to a continuous form, we first let $\mathbf{X}_n=\sum_{k=1}^{K}\mathbf{X}_{k,n}$, where
\begin{equation}
\mathbf{X}_{k,n+1}=\Tilde{A}_k \times_1 \mathbf{X}_{k,n} \times_2 U_k \times_3 Q_k + \frac{1}{K}\mathbf{X}_{0},    
\label{dp}
\end{equation}
by expanding Eq. \ref{dp} we can expand Eq. \ref{eq:di} as:
\begin{equation}
\label{eq:ex}
\begin{aligned}
\mathbf{X}_n &=\frac{K-1}{K}\sum_{k=1}^{K}\Tilde{A}_k^n \times_1 \mathbf{X}_0 \times_2 U_k^n \times_3 Q_k^n \\ &+ \frac{1}{K}\sum_{k=1}^{K}\sum_{i=0}^n \Tilde{A}_k^{i} \times_1 \mathbf{X}_0 \times_2 U_k^{i} \times_3 Q_k^{i}
\end{aligned}
\end{equation}

We replace the discrete $n$ with a continuous variable $t$ to extend Eq. \ref{eq:ex} to continuous form, which can be viewed as a Riemann sum from $0$ to $n$ on variable $i$. In practice, we clamp entries of $A_k$,$U_k$ and $Q_k$ in the interval $[0,1)$ to simplify Eq. \ref{eq:ex}. In this way, As $n$ goes to $\infty$, we have the integral formulation:
\begin{equation}
\label{eq:eq}
      \mathbf{X}(t)= \frac{1}{K}\sum_{k=1}^{K}\int_0^{t+1}\Tilde{A}_k^s \times_1 \mathbf{X}_0 \times_2 U_k^{s} \times_3 Q_k^{s}ds.
\end{equation}
\textbf{Proposition 1}: \textit{The first-order derivative of $X(t)$ in Eq. \ref{eq:eq} can be formulated as follows:}
\begin{equation}
    \label{eq:pr1}
    \begin{aligned}
    \frac{d\mathbf{X}(t)}{dt}=& \mathbf{X}_0 + \frac{1}{K}\sum_{k=1}^{K}(ln(\Tilde{A}_k) \times_1 \mathbf{X}(t) + ln(U_k) \times_2 \mathbf{X}(t) \\
    &+ ln(Q_k) \times_3 \mathbf{X}(t))
\end{aligned} \end{equation}
\begin{proof} We directly calculate the first derivative of $\mathbf{X}(t)$ as:
\begin{equation}
\label{eq:const1}
    \frac{d\mathbf{X}(t)}{dt} = \frac{1}{K}\sum_{k=1}^{K}\Tilde{A}_k^{t+1} \times_1 \mathbf{X}_0 \times_2 U_k^{t+1} \times_3 Q_k^{t+1}
\end{equation}

To reduce computation cost, we consider the second derivative of $\mathbf{X}(t)$ and integrate over $t$ on both sides of the second-order differential equation. We get:
\begin{equation}
\label{eq:const2}
\begin{aligned}
    \frac{d\mathbf{X}(t)}{dt} &= \frac{1}{K}\sum_{k=1}^{K}(ln(\Tilde{A}_k) \times_1 \mathbf{X}(t) + ln(U_k) \times_2 \mathbf{X}(t) \\ &+ ln(Q_k) \times_3 \mathbf{X}(t)) + \textit{const}.
\end{aligned}
\end{equation}

To solve the const, we combine Eq. \ref{eq:eq}, Eq. \ref{eq:const1} and Eq. \ref{eq:const2}, and let $t \rightarrow -1$. We get:
\begin{equation}
\begin{aligned}
    const =\mathbf{X}_0 - [ln(\Tilde{A}_k) \times_1 \mathbf{X}(-1) + ln(U_k) \\ \times_2 \mathbf{X}(-1) + ln(Q_k) \times_3 \mathbf{X}(-1)]=\mathbf{X}_0.
\end{aligned}
\end{equation}\end{proof}

The formulation of Eq. \ref{eq:pr1} can be further solved by an ODE solver such as the Runge-Kutta method or Euler method:
\begin{equation}
    \mathbf{X}(t) = \text{ODESolver}(\frac{d\mathbf{X}(t)}{dt},\mathbf{X}_0,t),
    \label{ode}
\end{equation}
which allows us to build it just as a block within the entire neural network. 
\subsection{The Temporal Module}\label{sec:tem_mo}
\textbf{1) Construction of Temporal Hypergraph}: We define $G_{te}=(V,\xi_{te},H_{te})$ as a $r$-uniform temporal hypergraph, where $V$ represents traffic sensors, and each hyperedge contains $r$ nodes with strong similarity. The binary incidence matrix $H_{te}$ encodes node-hyperedge relationships.

To measure the similarity between time series $X_i=(x_1,\dots,x_T)$ and $Y_j=(y_1,\dots,y_S)$ associated with nodes $v_i$ and $v_j$ respectively, we can use the Dynamic Time Warping (DTW) algorithm. In our implementation, we utilize the entire length of the training data for time series $X_i$ and $Y_j$ to measure the similarity between node $v_i$ and $v_j$.

\textbf{2) Hyperedge Evolving ODE}: Given that each hyperedge in the temporal hypergraph comprises nodes with the most similar timing patterns, it is intuitive to prioritize information propagation among these interconnected nodes. To achieve this efficiently, we employ hypergraph convolution as our information propagation scheme:
\begin{equation}
\label {eq:se}
\textbf{X}_{n+1} = A_{te} \times_1 \mathbf{X}_{n} \times_2 U \times_3 Q + \textbf{X}_0,
\end{equation}
where $A_{te}=D_{te}^{-1/2}H_{te}WB_{te}^{-1}H_{te}^T D_{te}^{-1/2}$ denotes the temporal hypergraph transform matrix, $U \in \mathrm{R}^{T \times T}$ and $Q \in \mathrm{R}^{F \times F}$ are two trainable weight matrix. Similar to Eq. \ref{eq:eq}, we can extend Eq. \ref{eq:se} to continuous form and we have the integral formulation:
\begin{equation}
\label{eq:se2}
      \mathbf{X}(t)= \int_0^{t+1}A_{te}^s \times_1 \mathbf{X}_0 \times_2 U^{s} \times_3 Q^{s}ds.
\end{equation}

Based on Proposition 1, we can derive the following corollary:

\textbf{Corollary 1}: \textit{The first-order derivative of X(t) in Eq. \ref{eq:se2} can be formulated as following ODE:}
\begin{equation}
    \label{eq:pr2}
    \begin{aligned}
    \frac{d\mathbf{X}(t)}{dt}=& \mathbf{X}_0 + ln(A_{te}) \times_1 \mathbf{X}(t) + ln(U) \times_2 \mathbf{X}(t) \\
    &+ ln(Q) \times_3 \mathbf{X}(t))
\end{aligned}
\end{equation}

The formulation of Eq. \ref{eq:pr2} can be solved using a designated ODE solver in the form of Eq. \ref{ode}.

\subsection{Others}
Inspired by Graph WaveNet\cite{graphwavenet}, we use a 1-D dilated causal temporal convolution as the temporal convolution layer (TCN), which enlarges the receptive field and enables parallel computation, and addresses the gradient explosion problem. The dilated causal convolution operation is denoted as:
\begin{equation}
    \mathbf{X}\ast \mathbf{f}_{1\times k}(t) = \sum_{s=0}^{k-1}\mathbf{f}_{1 \times k}(s)\mathbf{X}(t-d\times s),
\end{equation}
where $\mathbf{X} \in \mathbb{R}^{T \times F}$ is the input of TCN, and $\mathbf{f}_{1\times k} \in \mathbb{R}^k$ is the convolution filter, and $d$ denotes the dilation factor. 

Given the ground truth traffic signal $Y \in (0,1)$, the objective function of traffic forecasting is evaluated via Huber loss\cite{huber1992robust} defined as:
\begin{equation}
\begin{aligned}
    L(Y,\hat{Y}) = \begin{cases}
        \frac{1}{2}(Y-\hat{Y})^2&, \|Y-\hat{Y}\| \leq \delta\\
        \delta \|Y-\hat{Y}\| - \frac{1}{2}\delta&, \text{otherwise}
    \end{cases}
\end{aligned}
\end{equation}

where $\delta$ is a hyperparameter that controls the sensitivity to outliers. Huber loss combines the best properties of both quadratic and linear loss functions, allowing it to handle both small and large errors appropriately. 
The output of the stacked STHODE layers is aggregated using a max-pooling layer, which selectively combines the information from different blocks. The aggregated features are then passed into a two-layer MLP, which further transforms the features into the final predictions. This overall architecture enables the model to effectively capture both spatio-temporal dependencies.

\begin{table*}[htbp]
\caption{Traffic forecasting on PeMS03,PeMS04,PeMS07 and PeMS08}
\begin{center}
\begin{tabular}{c c c c c c c c c c c c c}
\hline
\multirow{2}{*}{\textbf{Model}} & \multicolumn{3}{c}{\textbf{PeMS03}} & \multicolumn{3}{c}{\textbf{PeMS04}} &\multicolumn{3}{c}{\textbf{PeMS07}}&\multicolumn{3}{c}{\textbf{PeMS08}}\\
\cline{2-13}
&MAE&RMSE&MAPE&MAE&RMSE&MAPE&MAE&RMSE&MAPE&MAE&RMSE&MAPE\\
\hline
\hline
STGCN &17.55& 30.42 &17.34\%& 21.16 &34.89&13.83\% &25.33 &39.34& 11.21\%& 17.50& 27.09&11.29\%\\
DCRNN &17.99 &30.31& 18.34\% &21.22 &33.44&14.17\% &25.22& 38.61& 11.82\% &16.82 &26.36 &10.92\%\\
GraphWaveNet &19.12 &32.77& 18.89\% &24.89 &39.66 &17.29\% &26.39 &41.50& 11.97\% &18.28 &30.05 &12.15\%\\
ASTGCN(r) &17.34& 29.56& 17.21\% &22.93 &35.22& 16.56\% &24.01& 37.87& 10.73\% &18.25 &28.06 &11.64\%\\
STSGCN & 17.48 &29.21 &16.78\% &21.19 &33.65 &13.90\% &24.26 &39.03& 10.21\% &17.13 &26.80& 10.96\%\\
STFGNN &16.77 &28.34 &\underline{16.30}\% &20.48 &32.51 &16.77\% &23.46 &36.60& \textbf{9.21\%} &16.94 &26.25& \underline{10.60\%}\\
STGODE &16.50& 27.84& 16.69\% &20.84& 32.82& 13.77\% &22.59& 37.54 &10.14\% &16.81& 25.97& 10.62\%\\
STHODE & \textbf{15.51}&\textbf{26.16} & \textbf{15.88\%}& \textbf{19.61}&\textbf{30.97} &\textbf{13.45\%} & \textbf{21.72}& \textbf{34.63} & \underline{9.82\%}&\textbf{15.43}& \textbf{24.39} & \textbf{10.27\%}\\
\hline
\end{tabular}
\label{tab2}
\end{center}
\end{table*}

\section{Experiments}
This section introduces the experiments on four real-world highway datasets to investigate the effectiveness and robustness of the proposed STHODE. The following research questions are answered:\\
\textbf{RQ1:} How does STHODE perform in traffic forecasting compared to the baseline models?\\
\textbf{RQ2:} How does each component of the model contribute to the performance of our solutions?\\
\textbf{RQ3:} How do different hyper-parameters influence the model performance?

\subsection{Experimental Setting}
\textbf{Datasets:} All the datasets are collected by the Caltrans Performance Measurement System(PeMS) in real-time 30 seconds, including four traffic flow datasets sampled from different districts or different periods. Following\cite{DCRNN}, we set the sample rate to 5 minutes and apply Z-Score normalization to inputs.

\textbf{Evaluation Metrics:} Mean Absolute Error (MAE), Root Mean Squared Error (RMSE), and Mean Absolute Percentage Error (MAPE). 

\textbf{Baselines for Comparison:} Various baselines are compared with the proposed STHODE: TCN\cite{bai2018empirical}, STGCN\cite{STGCN}, DCRNN\cite{DCRNN}, Graph WaveNet\cite{graphwavenet}, ASTGCN\cite{guo2019attention}, STSGCN\cite{STSGCN}, STFGNN\cite{li2022spatial}, STGODE\cite{STGODE}.

\textbf{Parameters Settings:} All experiments are implemented by Pytorch 2.0.0 on NVIDIA GeForce RTX 3090 GPU. We split all datasets with a ratio of 6:2:2 into training sets, validation sets, and test sets. One hour of historical data is used to predict traffic conditions in the next hour. We use Adam as our optimizer and set the learning rate to 0.001. The batch size is 16 and the training epoch is 200. The temporal convolution block has hidden dimensions of 64,32,64.

\subsection{Performance Comparison(RQ1)}
Table \ref{tab2} displays the performance comparison results of our STHODE method with various baseline approaches for traffic forecasting. Overall, our proposed STHODE achieves the most competitive performance on the three metrics and significantly surpasses all baselines on all the datasets. The improvement in performance can be attributed to some key factors: 1) The spatial module enables STHODE to effectively model the road network topology, surpassing graph-based methods; 2) The temporal module enables STHODE to fully extract high-order temporal dependencies, further improving prediction accuracy. Leveraging the hypergraph structure enhances the representation of complex relationships between traffic sensors and road segments, resulting in superior prediction accuracy.

\subsection{Model Ablation and Effectiveness Analyses(RQ2)}
To analyze STHODE's components, we conducted ablation experiments on PeMS04: \textit{w/o spatial}: Removes the spatial hypergraph module. \textit{w/o temporal}: Removes the temporal hypergraph module. \textit{w/o ode}: Replaces the ODE solver with a hypergraph convolution layer. \textit{w/o adaptive}: Removes the adaptive hypergraph matrix from spatial hypergraph construction.

\begin{figure}[htbp]

  \centering
  \subfigure{\includegraphics[width=0.45\columnwidth]{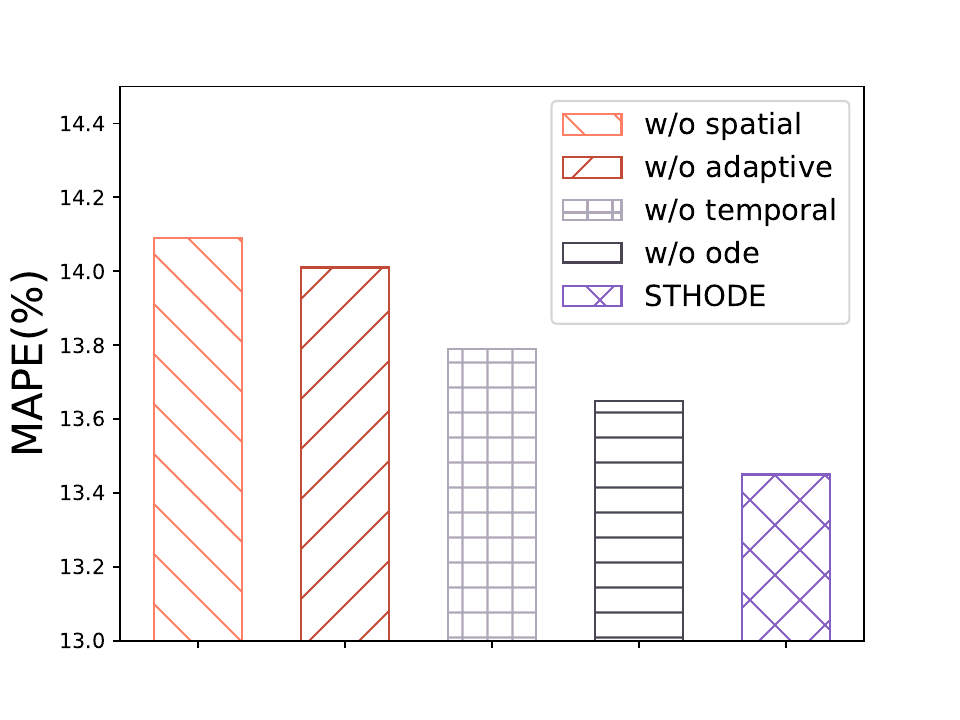}}
  \subfigure{\includegraphics[width=0.45\columnwidth]{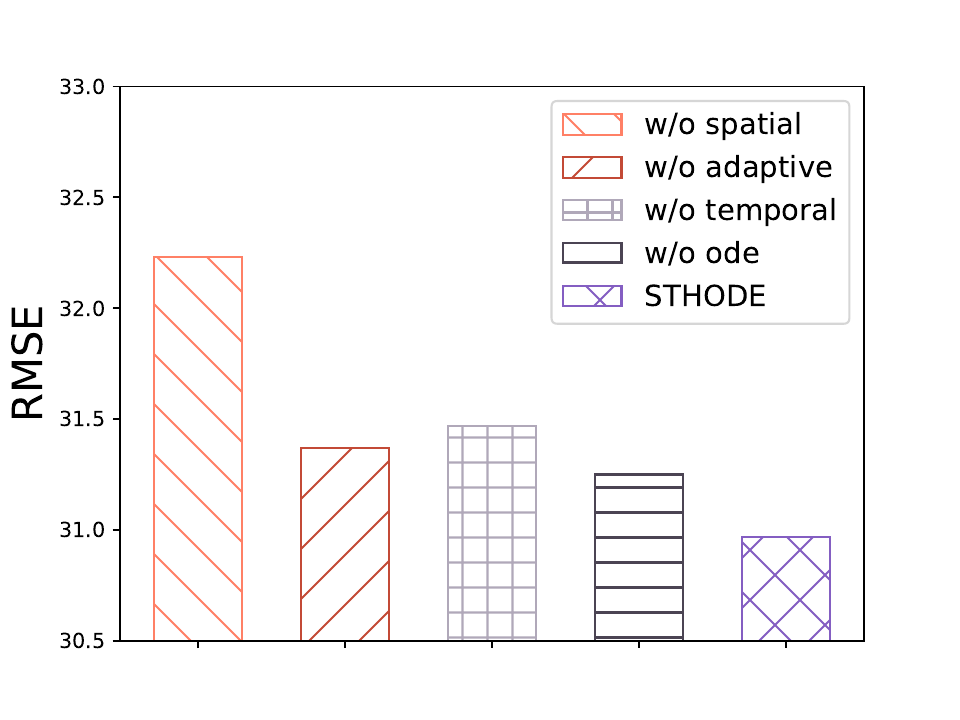}}
  \caption{Ablation experiments of STHODE on PeMS04 dataset.}
  \label{fig:ab}
\end{figure}

Results in Figure \ref{fig:ab} indicate: STHODE consistently outperforms all variants, emphasizing the importance of its components. Removing ODE layers (\textit{w/o ode}) reduces performance, which highlights their role in capturing data dynamics. Omitting either the spatial (\textit{w/o spatial}) or temporal module (\textit{w/o temporal}) results in performance decline, which is the significance of road network influences. \textit{w/o adaptive} highlights the importance of the adaptive hyperedge matrix in capturing data relationships and correlations.

\subsection{Hyperparameter Studies(RQ3)}
\begin{figure}[htbp]

    \centering
    \subfigure{\includegraphics[height=3cm]{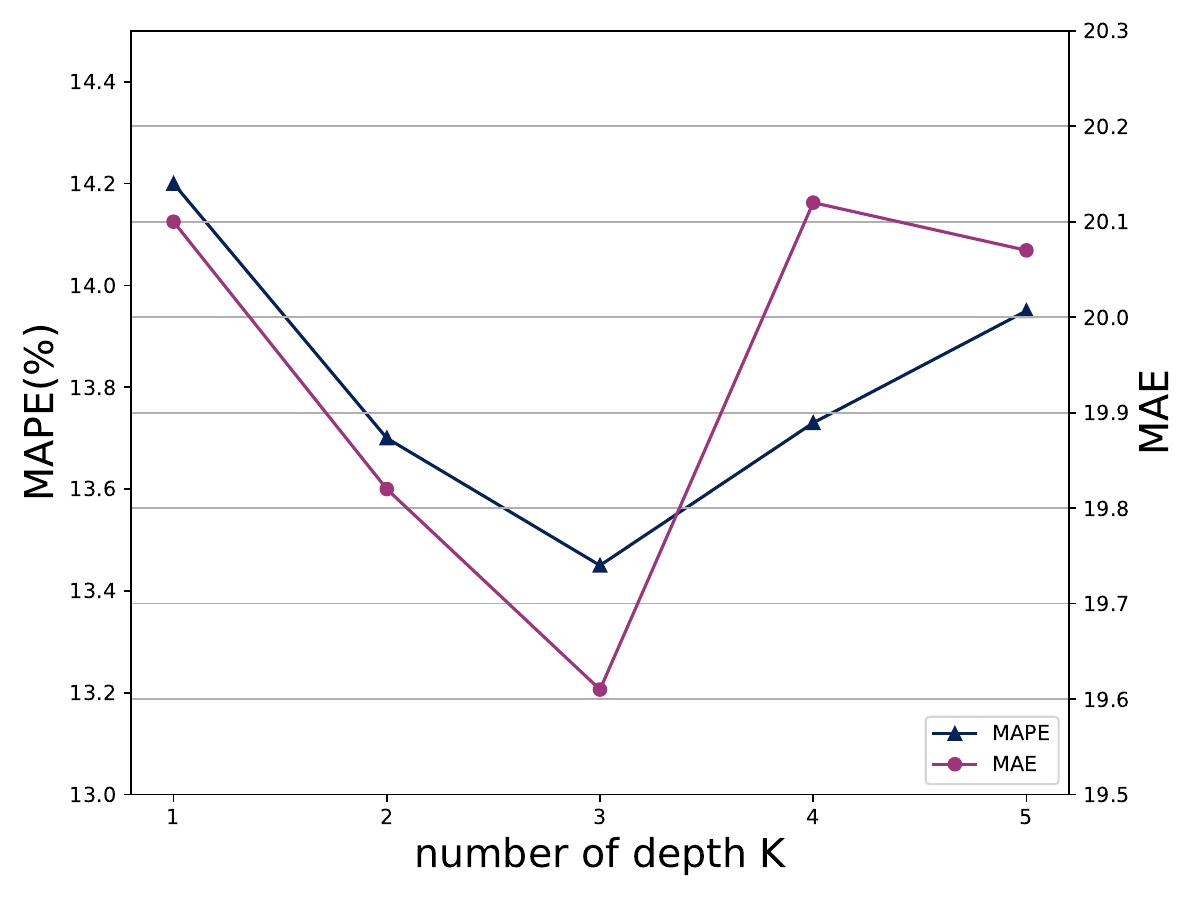}}
    \subfigure{\includegraphics[height=3cm]{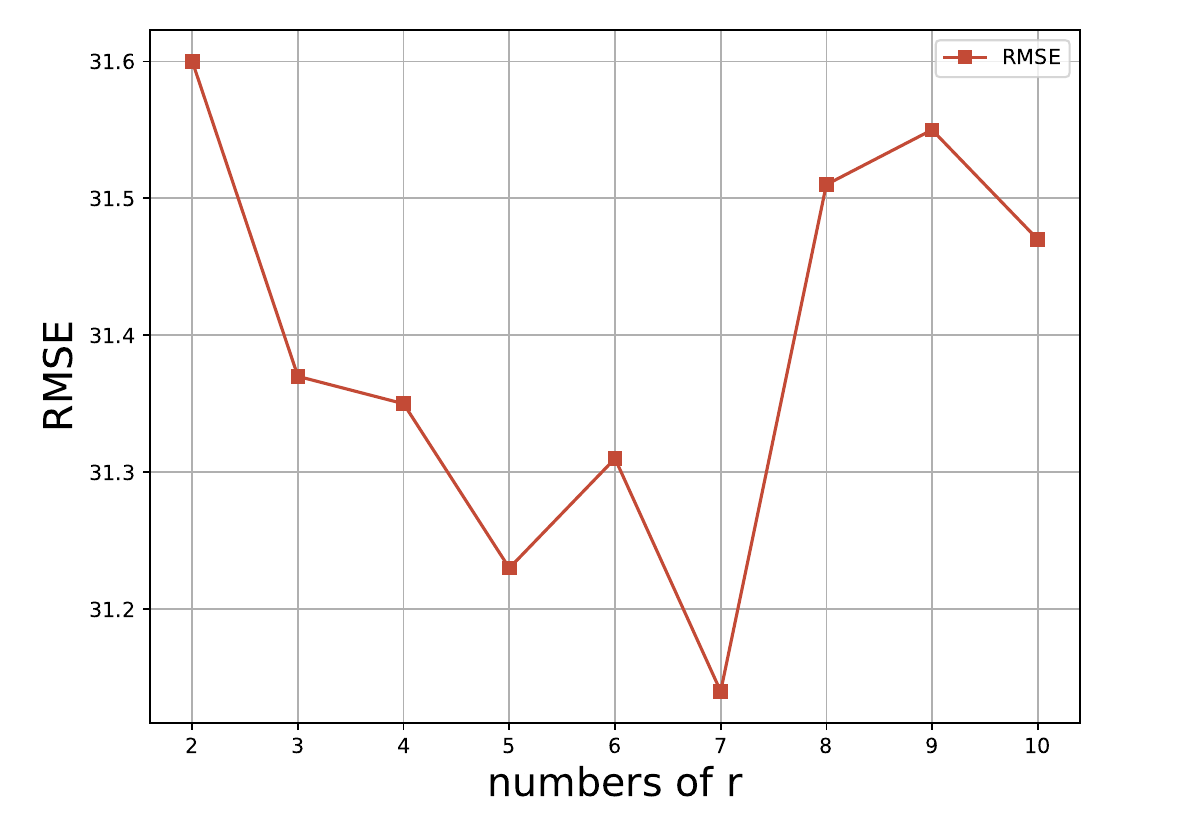}}
    % \subfigure{\includegraphics[width=0.45\columnwidth]{par1.pdf}}
    % \subfigure{\includegraphics[width=0.45\columnwidth]{par2-2.pdf}}
    \caption{Performance comparison w.r.t. different number of depth $K$ for spatial hypergraph and different $r$ for r-uniform temporal hypergraph}
    \label{fig:par}
\end{figure}

In Figure \ref{fig:par}, we present results from experiments on the PeMS04 dataset where we varied hyperparameters within the spatial and temporal hypergraph modules. Our observations are as follows:

Sensitivity of Spatial Module: Increasing $K$ from 1 to 3 generally improves performance. Notably, when $K=1$, the MixHop scheme reduces to the original hypergraph convolution scheme, confirming the effectiveness of MixHop. However, further increasing depth may lead to diminishing returns or overfitting.

Sensitivity of Temporal Module: Increasing $r$ from 2 to 7 generally improves performance, indicating effective capture of high-order temporal dependencies. However, further increasing $r$ may introduce noise or irrelevant information, leading to decreased performance.

\subsection{Case Study}
\begin{figure}[htbp]
    \centering
    \subfigure[]{\includegraphics[width=0.45\columnwidth]{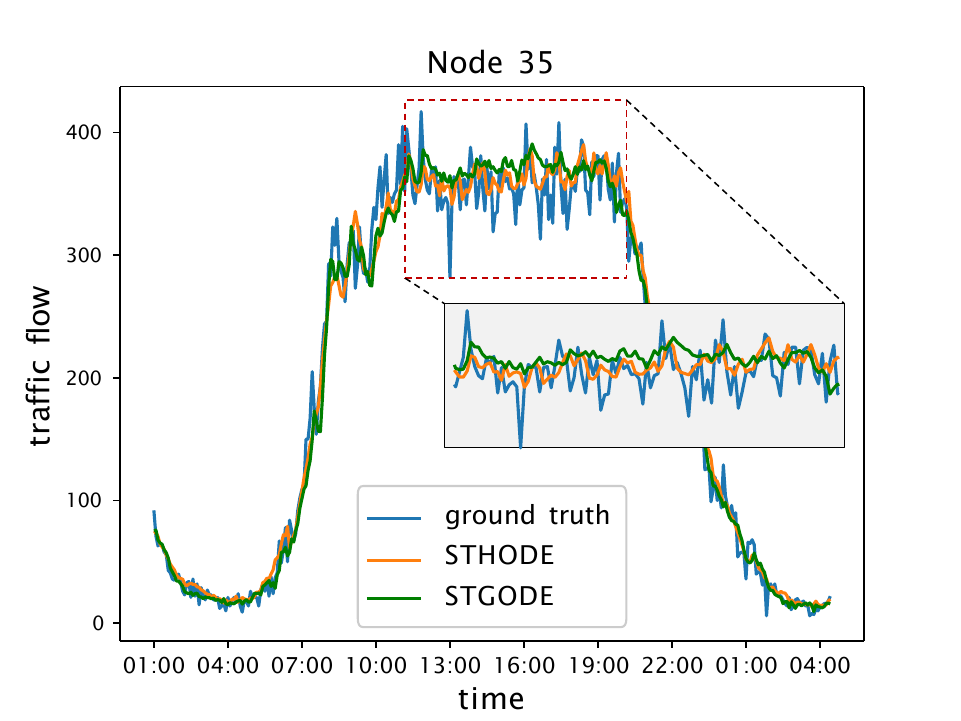}}
    \subfigure[]{\includegraphics[width=0.45\columnwidth]{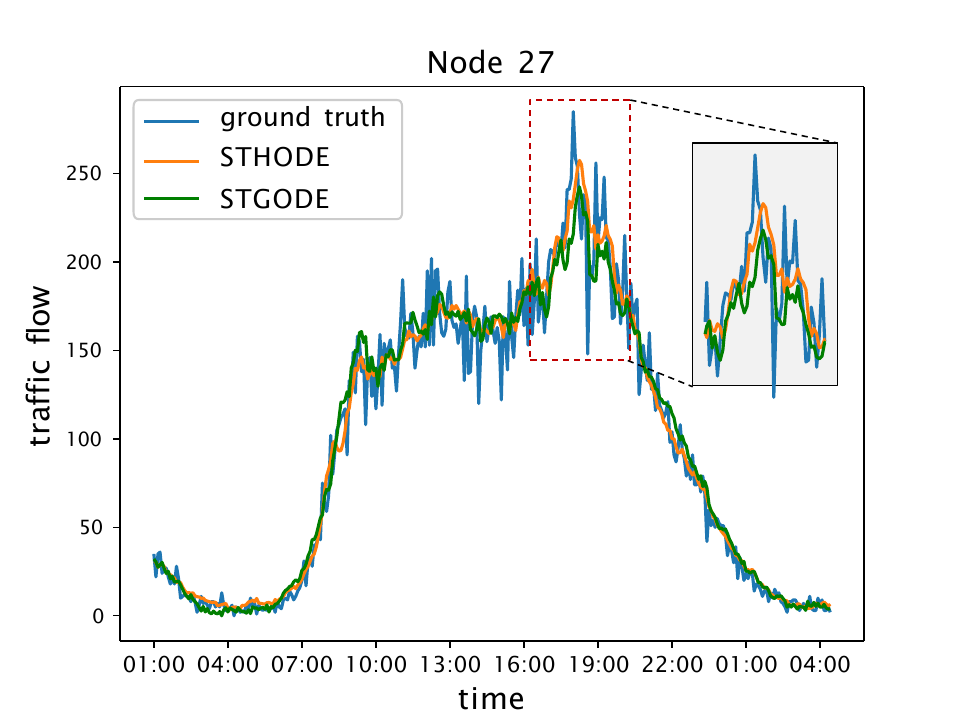}}
    \caption{The visualization of prediction results between our model(STHODE) and STODE.}
    \label{fig:case}
\end{figure}

A case study conducted on node 27 and node 35 from the PeMS04 dataset offers a detailed analysis of the performance of STHODE. In Figure \ref{fig:case}(a), during the heavy traffic flow from 10:00 to 19:00, STHODE consistently outperforms STODE. In Figure \ref{fig:case}(b), around 19:00, an abrupt change occurs and STHODE quickly adapts while maintaining high prediction accuracy. The ability to model the road network topology helps STHODE capture the correlation among different road segments, leading to improved prediction accuracy.

\section{Conclusion}
This paper introduces Spatio-Temporal Hypergraph ODE (STHODE). It models the road network topology and traffic dynamics to capture high-order spatio-temporal dependencies for short-term traffic predictions. STHODE utilizes two ODE-based modules that encode a spatial hypergraph and a temporal hypergraph working in parallel to capture high-order spatio-temporal dependencies respectively. Extensive experiments prove the effectiveness of STHODE over various existing methods.

\bibliographystyle{ieeetr}
\bibliography{reference}

\end{document}